\documentclass[letterpaper, 10 pt, conference]{ieeeconf}  
\pdfoutput=1

\IEEEoverridecommandlockouts                              

\usepackage{graphics} 
\usepackage{subfigure}
\usepackage{subcaption}
\usepackage{graphicx}
\usepackage{epstopdf}

\usepackage{epsfig} 
\usepackage{amsmath} 
\usepackage[T1]{fontenc}
\usepackage{bm}
\usepackage{cite}
\usepackage{subcaption}
\usepackage{booktabs}
\usepackage{multirow}
\usepackage{diagbox}
\usepackage{threeparttable}

\usepackage{tkz-euclide}
\usepackage{amsmath,amsfonts}
\usepackage{array}
\usepackage[caption=false,font=normalsize,labelfont=sf,textfont=sf]{subfig}

\usepackage{textcomp}
\usepackage{stfloats}
\usepackage{url}
\usepackage{verbatim}
\usepackage{graphicx}
\usepackage{cite}
\usepackage{times}
\usepackage{epsfig}
\usepackage{float}
\usepackage{wrapfig}
\usepackage{algorithm,algorithmicx,algpseudocode}
\usepackage{bm,xspace}
\usepackage{comment}
\usepackage{multirow}
\usepackage{balance}
\usepackage{booktabs}
\usepackage{etoolbox,siunitx}
\usepackage{calc}
\usepackage{pifont,hologo}
\usepackage{nicefrac}
\usepackage{makecell}
\usepackage{adjustbox}
\usepackage{threeparttable}
\usepackage{overpic}                
\usepackage{color}
\definecolor{abstractbg}{rgb}{0.89804,0.94510,0.83137}
\title{\LARGE \bf An Onboard Framework for Staircases Modeling Based on Point Clouds}

\author{Chun Qing*, Rongxiang Zeng*, Xuan Wu, Yongliang Shi and Gan Ma$^{\dag}$ 
\thanks{*Equal contribution, $\dag$Corresponding author} 
\thanks{C. Qing, R. Zeng, X. Wu and G. Ma are with the Sino-German College of Intelligent Manufacturing, Shenzhen Technology University, China. Y. Shi are with the Institute for AI Industry Research (AIR), Tsinghua University, China. Corresponding author: Gan Ma(magan@sztu.edu.cn)}} 

\begin{document}
\maketitle
\thispagestyle{empty}
\pagestyle{empty}

\begin{abstract} 
The detection of traversable regions on staircases and the physical modeling constitutes pivotal aspects of the mobility of legged robots. This paper presents an onboard framework tailored to the detection of traversable regions and the modeling of physical attributes of staircases by point cloud data. To mitigate the influence of illumination variations and the overfitting due to the dataset diversity, a series of data augmentations are introduced to enhance the training of the fundamental network. A curvature suppression cross-entropy(CSCE) loss is proposed to reduce the ambiguity of prediction on the boundary between traversable and non-traversable regions. Moreover, a measurement correction based on the pose estimation of stairs is introduced to calibrate the output of raw modeling that is influenced by tilted perspectives. 
Lastly, we collect a dataset pertaining to staircases and introduce new evaluation criteria. 
Through a series of rigorous experiments conducted on this dataset, we substantiate the superior accuracy and generalization capabilities of our proposed method. Codes, models, and datasets will be available at https://github.com/szturobotics/Stair-detection-and-modeling-project. 
\end{abstract}

\section{Introduction}
In the domain of legged robotics, distinguished by the locomotion of robots across varied terrains, two prominent strategies emerge: perception-free and perception-aware techniques\cite{sustech}. The latter strategy harnesses sensory inputs to enhance traversal capabilities, a concern in intricate environments such as staircases, frequently encountered in urban and indoor settings.

The landscape of staircase detection and modeling research can be categorized into two principal categories: learning-free and learning-based methods. 
Traditional paradigms favor learning-free approaches in the image or point cloud, in which staircases inherently exhibit structural characteristics in their sensing data. They facilitate their identification through specialized algorithms, often predicated on the detection of adjacent lines\cite{r1}, \cite{r16}, \cite{Delmerico2013AscendingSM} or plane structures\cite{r10}, \cite{r16}, \cite{thomasiros}. They are often characterized by strong assumptions or particular geometric constraints, constraining their adaptability to diverse environments\cite{r16}, \cite{perez2015detection}. 
The learning-free methods tend to require manual tuning tailored to each unique environment. 
In contrast, learning-based approaches predominantly rely on visual-based methods for staircase detection rather than point cloud-based methods. 
Visual-based methods face inherent limitations, particularly during nighttime operations, and are also directly affected by the accuracy of depth prediction and motion blur. Those directly impact the precise detection and segmentation of stairs by legged robots\cite{8675676}. 
Point cloud-based learning methods for staircase detection and modeling still fall short of fully eliciting performance. The efficiency of learning-based methods largely depends on the diversity and completeness of their datasets; however, to our knowledge, it is regrettable that no open-source staircase datasets that have enough diversity and completeness are available.
All of those works lack navigating information, such as central points and normals of traversable regions, making it inconvenient for legged robot navigation. 

\begin{figure}  
    \centering
    \includegraphics[width=1\linewidth]{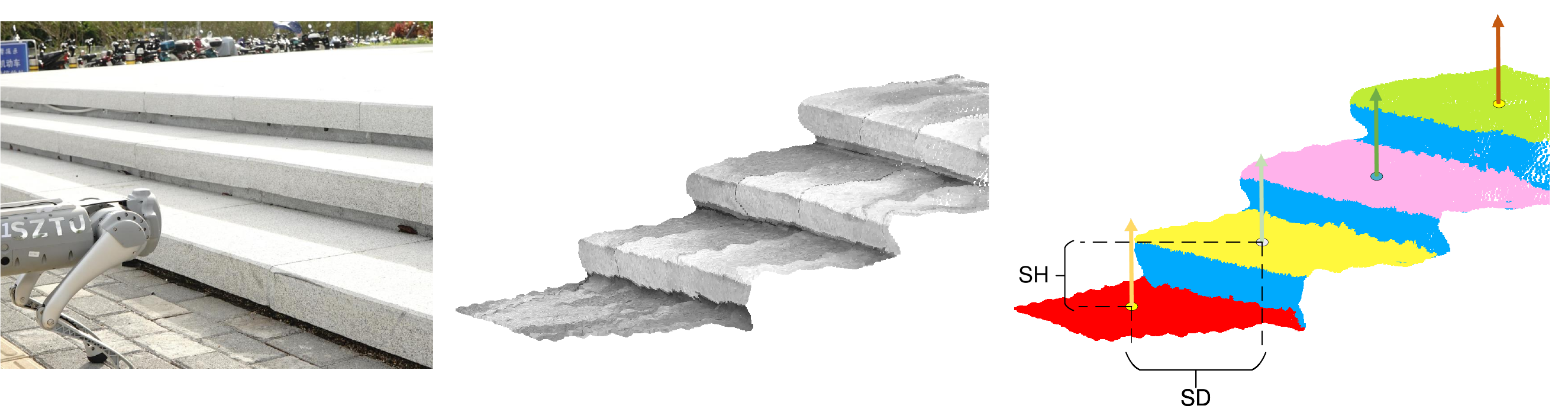}
    \caption{Staircase environment, staircase detection and modeling. The image illustrates the traversable regions predicted by our framework, with arrows indicating estimated normals. Ellipses positioned beneath the arrows represent center point estimations. The abbreviation 'SH' means step height, while the 'SD' refers to the step depth. The raw output of modeling is refined through measurement correction to derive stair parameters. Our method demonstrates robustness and accuracy across diverse lighting conditions and viewpoints, offering a reliable solution for robotic stair navigation in real-time.}
    \label{fig:1}
    \vspace{-4mm}
\end{figure}


To mitigate these limitations, we advocate for an onboard framework for staircase detection and modeling that integrates end-to-end instance segmentation with parameter modeling, as shown in Fig. \ref{fig:1}. 
This integrated approach yields a set of traversable regions and their corresponding parameters, encompassing navigation data. During training, we introduce a series of data augmentations to deal with overfitting due to the diversity of the dataset and prediction errors caused by illumination.
In addition, the CSCELoss to mitigate excessively aggressive positive predictions and the ambiguity at the boundaries of manual annotations. 
In the parameter modeling process, we derive the correction matrix from the step pose and perform the measurement corrections for step parameters. 
Furthermore, we also collect and label a dataset tailored to our requirements, emphasizing diversity through various light conditions and different viewing positions. We split the dataset for the usage of training, validation, and generalization. Ablation experiments for data augmentations, the CSCELoss, and the measurement correction are evaluated on the validation set and the generalization set. We select a well-done work\cite{r10} as a benchmark and complete a comparison test on both of these datasets.

The main contributions of our work are the following:
\begin{enumerate}
    \item[-] An end-to-end staircases instance segmentation and modeling framework that could handle real-time sensing data.
    \item[-] A series of data augmentations are applied to the geometric structures of point clouds to mitigate overfitting due to the dataset diversity and improve the prediction performance of partially corrupted point clouds on account of illumination interference.
    \item[-] The CSCELoss utilizes the constraints of modeled normal vectors to reduce the ambiguity in human-annotated regions of traversability and the fuzzy classification at boundaries during prediction.  
    \item[-] A measurement correction based on pose estimation is proposed for dealing with the parameter-modeling errors. 
    \item[-] A dataset for stair detection and modeling is collected and labeled within affiliated assessment criteria, and better performance has been demonstrated on our methods.
\end{enumerate}

\section{Related Works}
\subsection{Learning-free Methods}

Learning-free methods for staircase detection and modeling can be split into image-based and point cloud-based approaches. For image-based approaches, a Gabor filter was utilized in \cite{r1} to edge preservation, while a Hough transformation was applied in \cite{r8} to extract parallel lines of staircases from images. With the aid of point cloud, \cite{r16} assessed a scan-line grouping algorithm and a two-point random sampling algorithm to detect the risers of staircases. Additionally, Random Sample Consensus (RANSAC) and its derivative algorithms are pervasively used for segmenting planes of staircases, such as \cite{r10}, \cite{perez2015detection}, \cite{li2021real} and \cite{qian2014ncc}. In \cite{li2021real}, planes of descending staircases were estimated iteratively by RANSAC. In \cite{r10}, an algorithm combining the split and merge algorithm with RANSAC was donated as VoxelSAC, creating subsets of the point cloud by iteratively decreasing voxel size and then executing RANSAC on each generated subset. Moreover, for getting rid of plane segmentation, \cite{r11} delineated a conductive model for depicting parameters of staircases by segmented lines. In general, learning-free methods either manifest strong correlations with the environment or presuppositions regarding the geometric characteristics of the environment, necessitating extensive manual tuning for precise staircase detection.

\subsection{Learning-based Methods}
The utilization of deep learning in the identification of staircases has received limited attention, with the majority of studies concentrating on image-based staircase detection.
\cite{r12}  proposed a novel lightweight Fully Convolutional neural Network (FCN) architecture for the detection of staircases in weakly labeled natural images. In\cite{PRABAKARAN2023105844} a Deep Convolution Neural Network (DCNN) was utilized for the sTetro-D robot to detect a descending staircase. Furthermore,  YOLO\cite{redmon2016you} is also often used for stair inspections because of its accuracy and real-time performance\cite{8675676}\cite{9812456}\cite{jayawardana2022train}. Recently, deep neural networks are gradually employed for processing point cloud, particularly for segmentation, classification, and object reconstruction. For instance, \cite{r17} proposed a directional PointNet\cite{r18} model for classifying 3D point clouds of daily terrains, achieving a classification accuracy of 99\% for the testing set. On balance, these methods can only discover the staircases without realizing modeling and do not take the broken point clouds with high real time performance into consideration when detecting staircases.



\section{Methods} 
\subsection{System Overview:}
\begin{figure*}[htbp]
    \centering
    \includegraphics[scale=0.430]{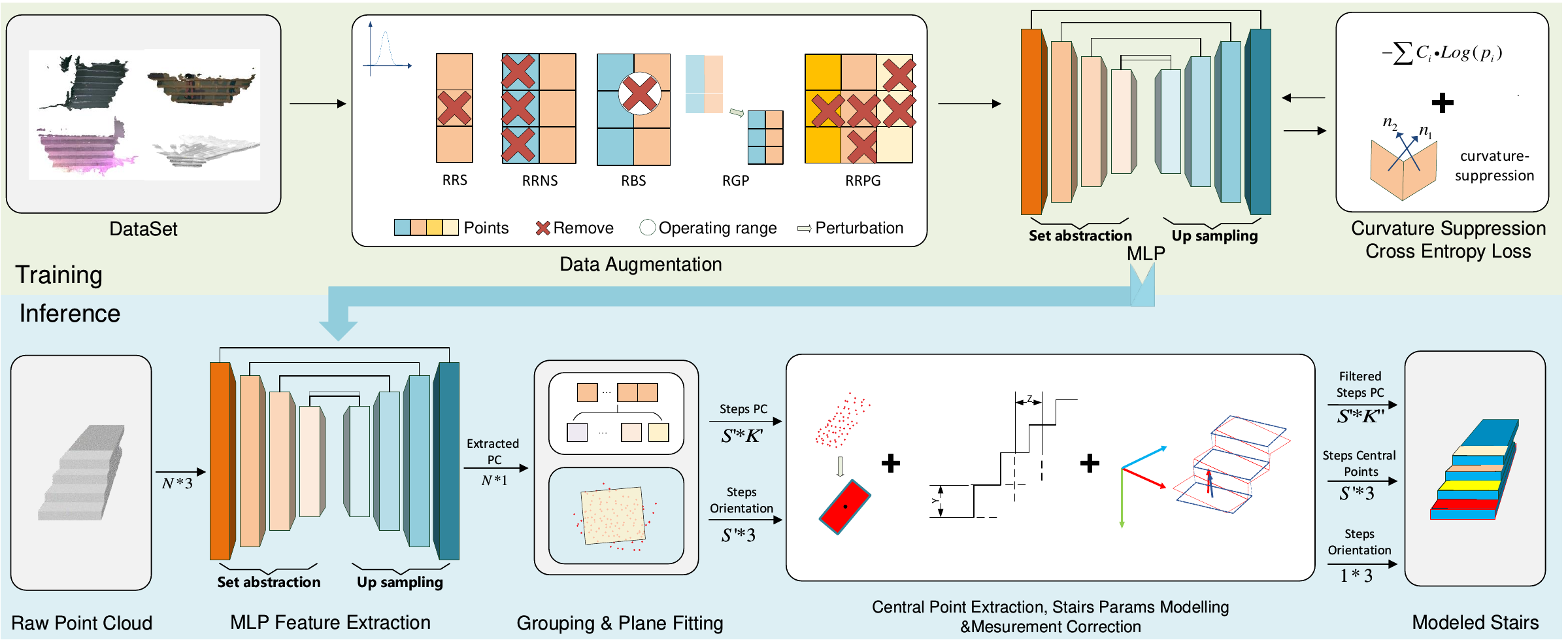}
    \caption{The training process of the predictor and the complete framework for staircase detection and modeling. Squares in the Data augmentation block represent points, red means steps and blue means non-steps. X marks represent the operation of removing. Arrows between blocks represent the flow of information, while the characters above are their dimensions and attributes (S denotes Step, K denotes points belonging to a specific S).}
    \label{fig:2}
\end{figure*}

The main framework is illustrated in Fig. \ref{fig:2}, which encompasses both the training procedure of the predictor and the complete process of staircase detection and modeling. In this paper, we define traversable regions as comprising predicted treads with geometry parameters(step central points, step normals, step height and depth) and navigation information(a list of central points and normals).
The framework incorporates PointNet++\cite{qi2017pointnet} as the backbone network for instance segmentation, resulting in an effective balance between performance and efficiency. We introduce five data augmentations and the CSCELoss to enhance the network's prediction performance. In the prediction and modeling process, the point cloud is input into the predictor downsampled, and outputs a set of step treads point clouds with order. Point cloud grouping is defined as an operation of making a set of predicted point clouds to initialize traversable regions. After grouping, the set of point clouds undergoes plane fitting to extract normals. Then use central point extraction and geometric relationships to produce central points and step height and depth for traversable regions. The current correction matrix is derived from the normals of treads or risers, if available, to correct the physical parameters before output traversable regions.

\subsection{Data Augmentation:} 

\textbf{Randomly Remove Steps (RRS):} It strategically omits varying numbers of stair treads from point clouds based on their labeling. Random values generated from a Gaussian distribution within the range of the staircase tread counts can be used for the point cloud extraction corresponding to the respective labels. Whether these values are used for the extraction depends on the probability of the RRS and can be adjusted using a threshold. This simulates the real-world scenario where sunlight induces mirrored reflections resulting in partial data loss. This augmentation is inclined towards removing stair treads located around the median stair treads count within the stair point clouds. 

\textbf{Randomly Remove Non-Steps (RRNS):} It takes effect in a way of probabilistic removing of non-step points from a given stair point cloud. A uniform distribution is used to determine the probability of enabling the operation. If in actuation, it removes stair risers points or non-step points from the point cloud. It simulates the appearance of industrial metal sheet stairs that lack risers. 

\textbf{Random Ball Sampler (RBS):} The sampler performs a fixed-size spherical region at central positions within the point cloud used to define a boundary for removing points that fall within it. After computing the distances between all points and the center of the spherical sampler, points are retained if their distance exceeds the radius. The sampler simulates the point cloud which has concave damage patterns due to specific angles of illumination. 

\begin{figure}
    \centering
    \includegraphics[width=0.8\linewidth]{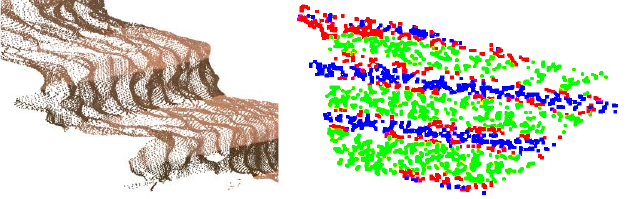}
    \caption{The phenomenons of point cloud undulations and overly aggressive stair treads predictions. Red represents FP and FN, blue represents TN and green represents TP in the right sub-image.}
    \label{fig:3}
    \vspace{-4mm}
\end{figure}

\textbf{Random Gaussian Perturbation (RGP):} Smooth surfaces of stairs captured by RGB-D cameras may display apparent fluctuations or undulations in the point cloud positions, as depicted in Fig. \ref{fig:3}. The lower depth precision of sensing exhibits notable variance in depth measurements. To address this issue, the augmentation employs a tiny Gaussian perturbation at each point to introduce additional undulations in the point cloud.

\textbf{Randomly Remove Points within Groups (RRPG):} RRPG compresses the number of points for different orders of steps in point clouds by multiplying them by a downsampling coefficient. It then employs a high-variance Gaussian function to generate the desired number of points for each category and subsequently performs random sampling on the resulting point cloud of steps.

\subsection{Curvature Suppression Cross Entropy Loss:} 
During the prediction, the network tends to make positive predictions, occasionally even for data points located on vertical risers. This behavior can largely be attributed to the inherent ambiguity observed during the manual annotation process, especially at the boundaries. As shown in Fig. \ref{fig:3}, the distribution of prediction errors primarily focuses on the upper part of the point cloud, with some minor inaccuracies at the intersection between the bottom treads and vertical risers. These inaccuracies could impact the calculation of the physical parameters of stairs. Given that the upper intersecting points are irrelevant to the robot's current foothold, it becomes crucial to impose higher penalties on points at lower intersections or on more pronounced vertical risers. Building on this concept, we formulated the CSCELoss function as Eq. \ref{eq:csce} and Eq. \ref{eq:2}, $cs$ refers to curvature suppression and $ce$ refers to cross-entropy. The $n$ represents the epoch threshold that triggers the $cs$ part of the loss when exceeds it.

\begin{equation}
\begin{aligned}
L_{csce} = 
 (1-k)\cdot L_{ce} + k\cdot (L_{cs} + L_{ce})
 \label{eq:csce}
\end{aligned}
\end{equation}
\vspace{-0.2cm}

\begin{equation}
\begin{aligned}
\scalebox{0.9}{$
k = \begin{cases}
    0, & \text{if epoch  $<$ $n$}\\
    1, & \text{if epoch  $\geq$ $n$}
\end{cases}$}
 \label{eq:2}
\end{aligned}
\end{equation}

This loss function combines the cosine distance between the normal vectors of positively predicted points and the average ground truth normal vectors of stair treads with the cross-entropy loss function. The curvature suppression loss is formulated as Eq. \ref{eq:3}. The abbreviation $s$ means the total step counts in a stair, while the $p$ means points of a step. The $N$ represents normals.

\begin{equation}
\begin{aligned}
L_{cs} = 
\scalebox{0.9}{$ \sum_{i=1}^{s}\sum_{j=1}^{p} \left \{a \left (1 - \left ( \frac{N_{gt,i}\cdot N_{pred,j}}{\|N_{gt,i}\|\cdot \|N_{pred,j}\|}  \right )^2\right )\right\} $}   
\label{eq:3}
\end{aligned}
\end{equation}
\vspace{-0.2cm}
\begin{equation}
\begin{aligned}
\scalebox{0.9}{$
a = \begin{cases}
    0, & \text{if step normal not exist }\\
    1, & \text{if step normal exist}
\end{cases}$}
\end{aligned}
\end{equation}

This approach penalizes points with prediction errors not only based on their classification error but also on their geometry constraint. Small boundary errors introduced during manual labeling are also averaged when calculating the loss function, reducing the impact of labeling ambiguities. 


\subsection{Stair Detection and Modeling:}
Stair detection concludes once the point cloud prediction is complete. Subsequently, a grouping procedure initialize traversable regions. Following grouping, we employ a RANSAC-based plane fitting\cite{FISCHLER1987726} on each tread's point cloud, which yields the extracted normal vector to determine the central point's orientation.

The point cloud resulting from plane fitting undergoes mean calculation to estimate the central point's position. This calculation reduces computing time compared to plane segmentation methods which are used to derive step parameters. Additionally, it averages out the noise effects caused by outliers during prediction.

The step parameter modeling component relies on geometric relationships to derive the depth and height. In this aspect, certain assumptions are made. It is assumed that the predicted central points are accurate enough, and the stair treads to be solved for are parallel to the camera coordinate system's $Z$-axis and $Y$-axis. The formulas for calculating parameters refer to Eq. \ref{eq:stepdepth} and Eq. \ref{eq:stepheight}, which leverage the projections of distances between nearby central points onto the $Z$ and $Y$ axes of the camera coordinate system. The abbreviation $CP$ refers to a central point, $n$ refers to the next, and $p$ refers to the previous. The capital $N$ means step numbers between step $n$ and step $p$. 

\begin{equation}
\begin{aligned}
\scalebox{1}{$
Step_{d} = \frac{CP_{z,n} - CP_{z,p}}{N(n,p)}$}
\label{eq:stepdepth}
\end{aligned}
\end{equation}

\begin{equation}
\begin{aligned}
\scalebox{1}{$
Step_{h} = \frac{CP_{y,n} - CP_{y,p}}{N(n,p)}$}
\label{eq:stepheight}
\end{aligned}
\end{equation}

\begin{figure}[t]
    \centering
    \includegraphics[width=0.75\linewidth]{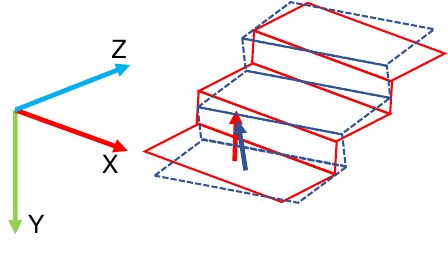}
    \caption{Demonstration of measurement correction: In the illustration, the red points represent the central points after correction, while the blue points represent the central points before correction. The rotation matrix "R" is computed in real time based on the current normal vector to achieve the rotation correction.}
    \vspace{-0.5cm}
    \label{fig:4}
\end{figure}

To enhance modeling accuracy, efforts are directed towards achieving the two prerequisites mentioned earlier. The accuracy of central point prediction is primarily improved through advancements in the network's performance. In scenarios involving multi-angle captures of stairs, deviations in the predicted central points with respect to the camera coordinates may occur. Therefore, the condition of parallelism between the subsequent treads and the coordinate axes is mainly addressed through rotation transformations applied to achieve alignment, referring to Fig. \ref{fig:4}. This is accomplished by solving the rotation matrix in a modeling loop that aligns the predicted normal vector of the nearest tread and riser with the target alignment axis, referring to Eq. \ref{eq:rotation1}. The $a_{xyz}$ represents the axis of camera coordinates and $n_{pred}$ means step normals. In the way of detecting the staircase has no risers, the rotation matrix is estimated using PCA \cite{shlens2014tutorial}. This derived transformation is then used to rotate the central point to be solved, resulting in a significant improvement in accuracy after this correction, as shown in Eq. \ref{eq:rotation2}. The $P$ represents points before correction and the $P_{correct}$ means points after correction. 

\vspace{0cm}

\begin{equation}
\begin{aligned}
\scalebox{0.97}{$
R_{t} = R_{t}(a_{xyz},n_{pred})$}
\label{eq:rotation1}
\end{aligned}
\end{equation}
\vspace{-0.5cm}
\begin{equation}
\begin{aligned}
\scalebox{0.97}{$
P_{correct} = R_{t}P$}
\label{eq:rotation2}
\end{aligned}
\end{equation}

\section{Experiments} 
The experimental setup involves evaluating proposed data augmentations, CSCELoss, measurement correction, and comparing them with related studies. For this purpose, we create a dataset and establish evaluation criteria. The training is conducted on a system equipped with an Intel i9-12900K CPU and an NVIDIA RTX 3080TI GPU with 12GB memory, running on the Win10. Inference is performed on a laptop featuring an Intel i5-11400H CPU @2.70GHz and an NVIDIA RTX 3050 GPU with 4GB memory. We implement our complete pipeline using open3d 0.17 \cite{DBLP:journals/corr/abs-1801-09847} and PyTorch 2.0.1 \cite{DBLP:journals/corr/abs-1912-01703}.


\subsection{Dataset}
We gathered and labeled a specialized point cloud dataset for staircase detection and modeling to enhance the specificity and diversity of various staircase datasets. Our dataset comprises nearly 800 depth images collected under various conditions, including strong and weak lighting, different viewpoints, and heights. The dataset encompasses 11 groups of staircase data. The data collection was conducted using an Intel RealSense D455 camera and stored in text format, organized as $XYZ-RGB-Label-Normal$. The labels of the dataset include traversable regions, step orders, step normals, and central points, also gathered the ground truth step parameters including height and depth. 

In the paper, the dataset is divided into equally distributed training and validation sets, with a ratio of 9:1. We additionally gathered a small dataset (5 classes, each with ten items) for generalization testing, which was never used during training and includes all the required attributes for our testing. We uniformly downsampled the dataset point clouds to contain around 10,000 points to facilitate training and usage.

\subsection{Metrics}
In addition to commonly used metrics for assessing the physical parameters of stairs, we introduced metrics specifically targeting central points and step normals.

\textbf{Depth Error and Height Error(DE, HE):} 
The mean absolute error (MAE) evaluates the predicted stairs parameters and measured parameters of 11 types of stairs. 

\textbf{Treads Quantities Error (FN, FP):}
We are interested in identifying traversable regions on stairs, paying close attention to the predicted tread counts in the camera view versus the ground truth. 'FN' denotes situations where predicted treads are less than ground truth, while 'FP' signifies the opposite, indicating incorrect predictions. 'FP' also includes cases where stairs are predicted in the wrong position.

\textbf{Central Point Error (CPE):}
The Euclidean distance evaluates the estimated central point position (the first step) and the ground truth central point position.

\textbf{Normal Orientation Error (NE):} 
The cosine distance compares an estimated normal of a stair tread and the ground truth normals of a step.

\begin{table*}[!t]
\caption{Ablation Study}
\centering
\begin{threeparttable}
\renewcommand\arraystretch{1.0}
\begin{tabular}{lccccccc}
\toprule 
\begin{tabular}[c]{@{}c@{}}\makecell[c]{Methods}\end{tabular} &
\begin{tabular}[c]{@{}c@{}} DE(m) \end{tabular}&
\begin{tabular}[c]{@{}c@{}} HE(m) \end{tabular}&
\begin{tabular}[c]{@{}c@{}} FP(\%) \end{tabular}&
\begin{tabular}[c]{@{}c@{}} FN(\%) \end{tabular}&
\begin{tabular}[c]{@{}c@{}} CPE(m) \end{tabular} &
\begin{tabular}[c]{@{}c@{}} NE(m) \end{tabular} &
\begin{tabular}[c]{@{}c@{}} TC(sec) \end{tabular} \\
\midrule 
Baseline&0.107         &0.02          &24.5         &5.1         &0.045         &0.997         &$\leq$0.2\\
NoCor-Baseline &0.11&0.101        &28.2&3.8&0.043     &0.997&$\leq$0.2\\
CSCEL   &\textbf{0.081}&0.021         &29.4         &\textbf{0}  &0.046         &0.997         &$\leq$0.2\\
RRS     &\textbf{0.074}&\textbf{0.017}&32.1         &\textbf{1.2}&\textbf{0.039}&0.996         &$\leq$0.2\\
RRNS    &\textbf{0.07} &0.02          &\textbf{20.5} &5.12         &0.045     &0.987&$\leq$0.2 \\
RBS     &\textbf{0.088}&0.022         &\textbf{22.8}&\textbf{3.8}&\textbf{0.044}&0.997         &$\leq$0.2\\
RGP     &\textbf{0.095}&0.022         &\textbf{19.2}&5.1         &0.051         &0.99          &$\leq$0.2\\
RRPG    &\textbf{0.083}&0.021         &\textbf{24.3}         &\textbf{0}  &0.045         &0.997  &$\leq$0.2\\
All     &\textbf{0.067}&0.021         &29.5  &\textbf{3.8}&\textbf{0.041}         &0.997         &$\leq$0.2\\
\midrule
Gen-Baseline &0.11         &0.038            &14  &44         &0.131          &0.985&$\leq$0.2\\
Gen-RRS &\textbf{0.081}    &0.041            &32  &\textbf{2} &\textbf{0.082} &0.978&$\leq$0.2\\
Gen-RRNS&\textbf{0.079}    &0.040            &48  &\textbf{12}&0.13           &0.982&$\leq$0.2\\
Gen-RBS &\textbf{0.0712}   &\textbf{0.024}   &34  &\textbf{24}&\textbf{0.11}  &\textbf{0.991}&$\leq$0.2\\
Gen-All &\textbf{0.06}     &\textbf{0.029}   &40  &\textbf{14}&\textbf{0.08}  &0.977&$\leq$0.2\\
\bottomrule
\end{tabular}
\begin{tablenotes}
        \footnotesize
        \item[*] The values in \textbf{bold} indicate that they exceed their baseline. 
        \item[**] The abbreviation 'NoCor' means No Correction, while the 'Gen' represents 'Generalization'. The 'TC' represents time-cost.
\end{tablenotes}
\end{threeparttable}
\label{tab:AB}
\vspace{-4mm}
\end{table*}

\subsection{Ablation Study}
We evaluate data augmentations, the CSCELoss, and the measurement correction on the validation set. We assess the generalization performance through metrics, thus discussing generalization tests alongside other evaluation criteria, and test the RRS, RRNS, RBS and a combined technique on the generalization set.
The outcomes of these experiments are presented in Table \ref{tab:AB}. The baseline is the same pipeline configuration without any training techniques. 'NoCor' means it runs without a measurement correction. The 'All' in the table represents the combination of all techniques listed. 

\begin{figure*}[t]
	\centering
	\subfigure[The raw image]{
		\begin{minipage}[t]{0.22\linewidth}
            \raggedright			
			\includegraphics[width=1.1\linewidth]{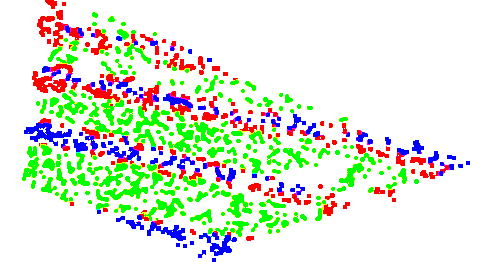}\\
			\vspace{-0.5cm}
		\end{minipage}%
	}%
	\subfigure[The CSCELoss]{
		\begin{minipage}[t]{0.22\linewidth}
			\raggedright
			\includegraphics[width=1.2\linewidth]{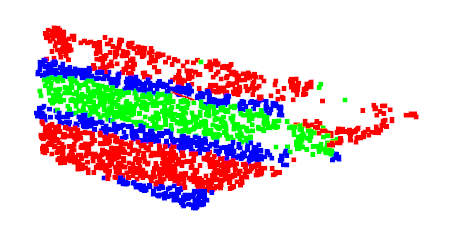}\\
			\vspace{-0.5cm}
		\end{minipage}%
	}%
	\subfigure[The correction of Z-axis]{
		\begin{minipage}[t]{0.22\linewidth}
			\raggedright
			\includegraphics[width=1.2\linewidth]{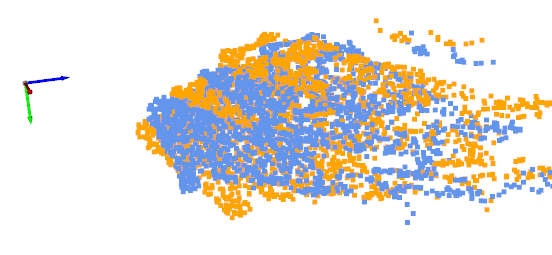}\\
			\vspace{-0.5cm}
		\end{minipage}%
	}%
 	\subfigure[The correction of Y-axis]{
		\begin{minipage}[t]{0.22\linewidth}
			\raggedright
			\includegraphics[width=1.2\linewidth]{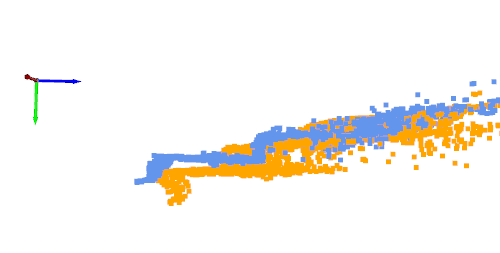}\\
			\vspace{-0.5cm}
		\end{minipage}%
	}%
	\centering
         \vspace{-0.1cm}
	\caption{The Visualized results of the impact of the CSCELoss and measurement correction. In sub-image (a), the points in red represent FP and FN after the prediction of baseline. The points in blue mean risers and green mean treads. In sub-image (b), the FP and FN on the risers nearly disappear. There we colored the treads in red and green, alternately. In sub-images (c) and (d), the points in orange represent before the correction, remaining points in cornflower blue represent after the correction.}
	\vspace{-0.5cm}
	\label{fig:ab}
\end{figure*}

In comparison to the baseline, all employed techniques exhibit efficacy. We provide visual results to demonstrate the impact of CSCELoss and measurement correction, as shown in Fig. \ref{fig:ab}. The combined strategy 'all' yielded the best performance in depth evaluation. The best performance has been achieved by our methods in terms of HE and NE. The CPE is acceptable relative to the tread's size, although there is still some room for improvement. The test runs on the validation set posted that the CSCELoss and RRPG can eliminate the false negative prediction to zero. The three augmentations and 'All' take effect in the generalization test, and our methods work to enhance the generalization of the framework. 

The DE and HE have two primary causes. One is the bias in the central points of steps, which affects the errors referred to in Eq. \ref{eq:stepdepth} and Eq. \ref{eq:stepheight}. This can be observed in the trends of the CPE, DE, and HE, which decrease when we apply training techniques. The other reason is measurement correction. Comparing the baseline and baseline without correction, the HE is mainly reduced by the correction. Here we prioritize the precision of height measurements over depth measurements in the stair modeling process. Depth measurement relies on the accurate segmentation of stair risers. Most stair risers are relatively narrow compared to the larger tread areas, making the precise prediction of their normals more challenging.

Regarding the reduction of FN to zero, we contend that the CSCELoss empowers the network to factor in normals of points, allowing for accurate predictions of the upper portions of stairs, which are challenging for manual annotation. This observation is further substantiated by the increase in FP. The RRPG helps mitigate overfitting in scenarios where point counts and distance exhibit proportionality, aiding the network in generating positive responses for sparse points.

\begin{figure*}[t]
	\centering
	\subfigure[The raw input]{
		\begin{minipage}[t]{0.15\linewidth}
            \raggedright			
			\includegraphics[width=1.2\linewidth]{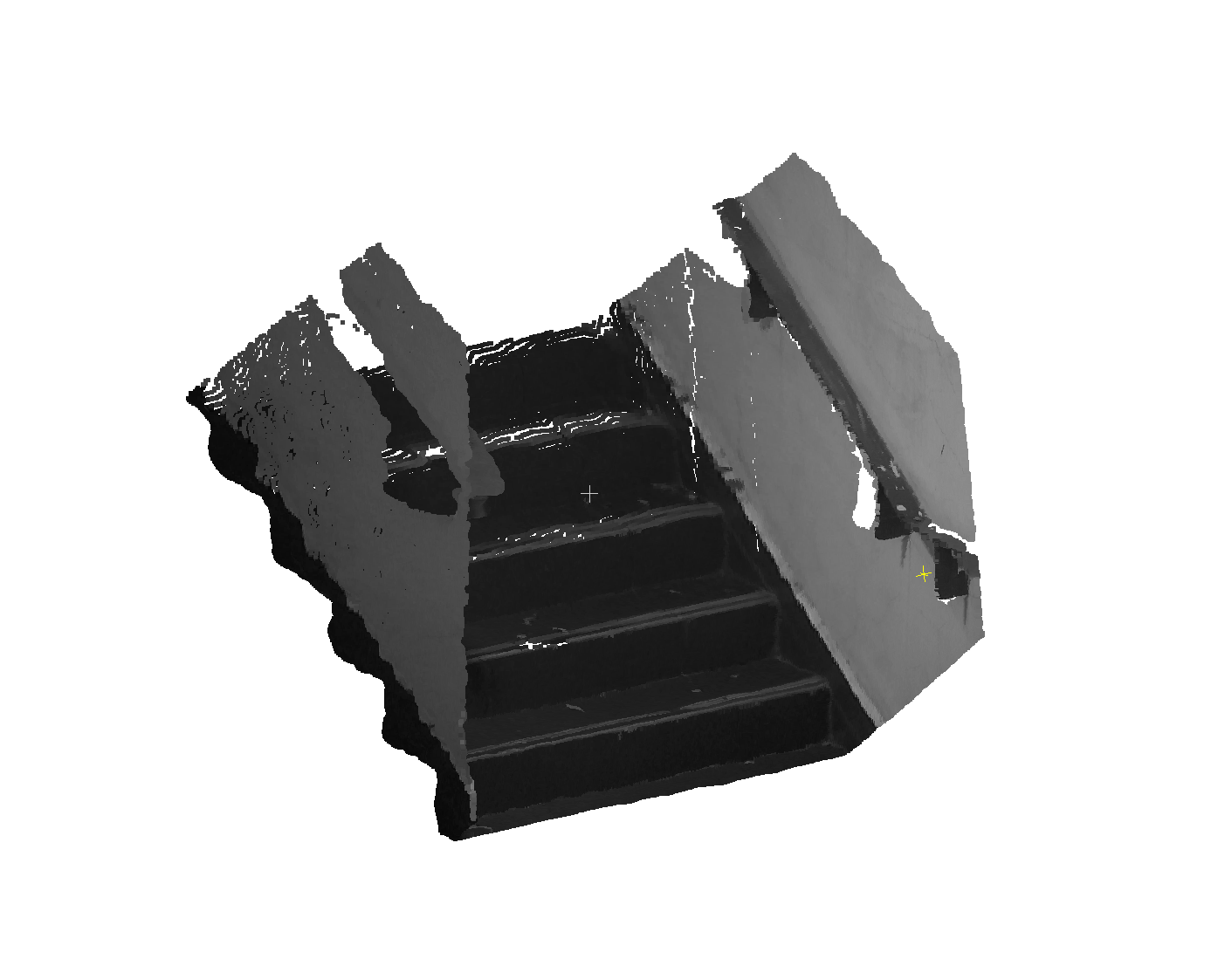}\\
			\vspace{0.02cm}
                \includegraphics[width=1.2\linewidth]{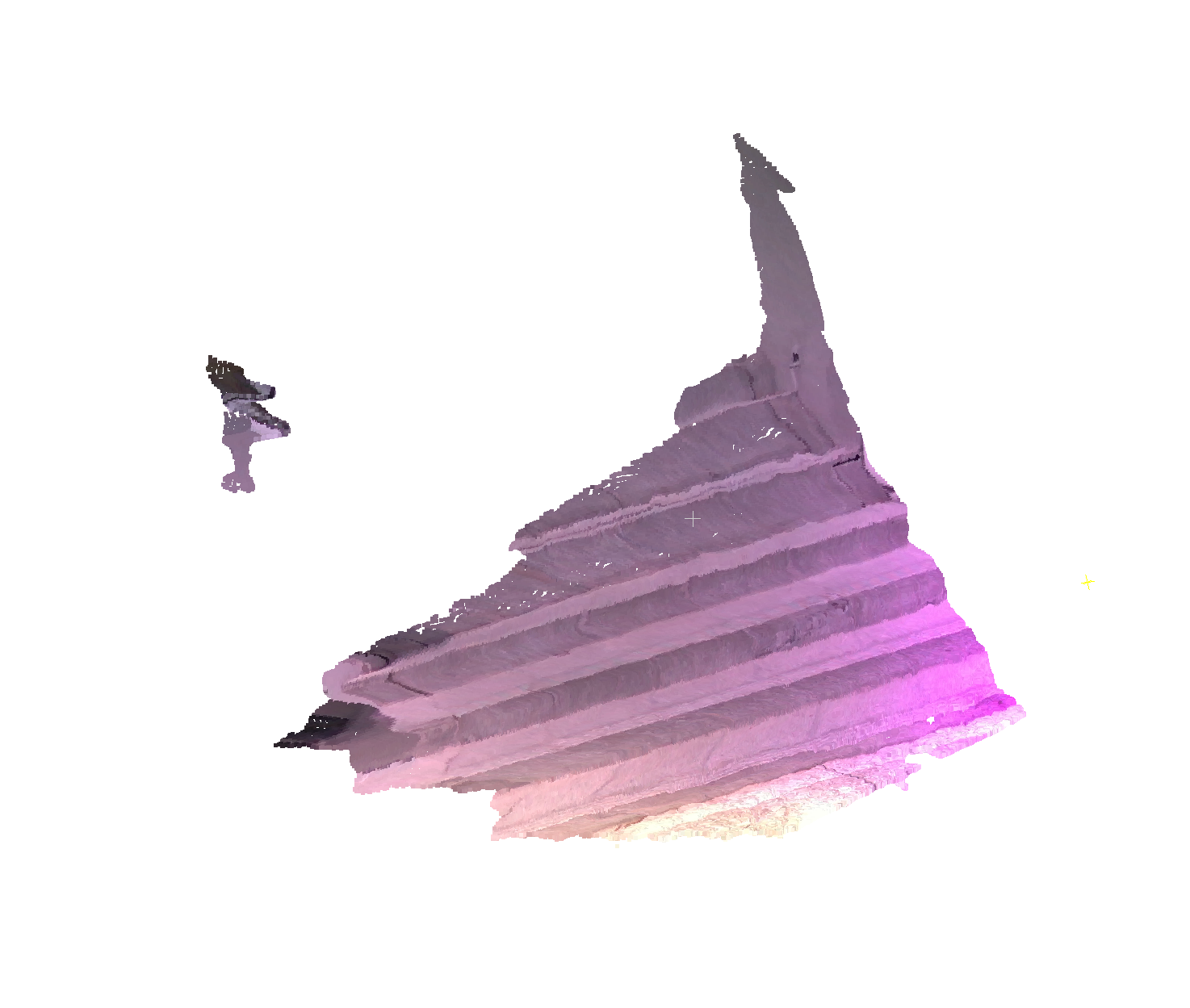}\\
			\vspace{0.04cm}
		\end{minipage}%
	}%
	\subfigure[Thomas]{
		\begin{minipage}[t]{0.15\linewidth}
			\raggedright
			\includegraphics[width=1.2\linewidth]{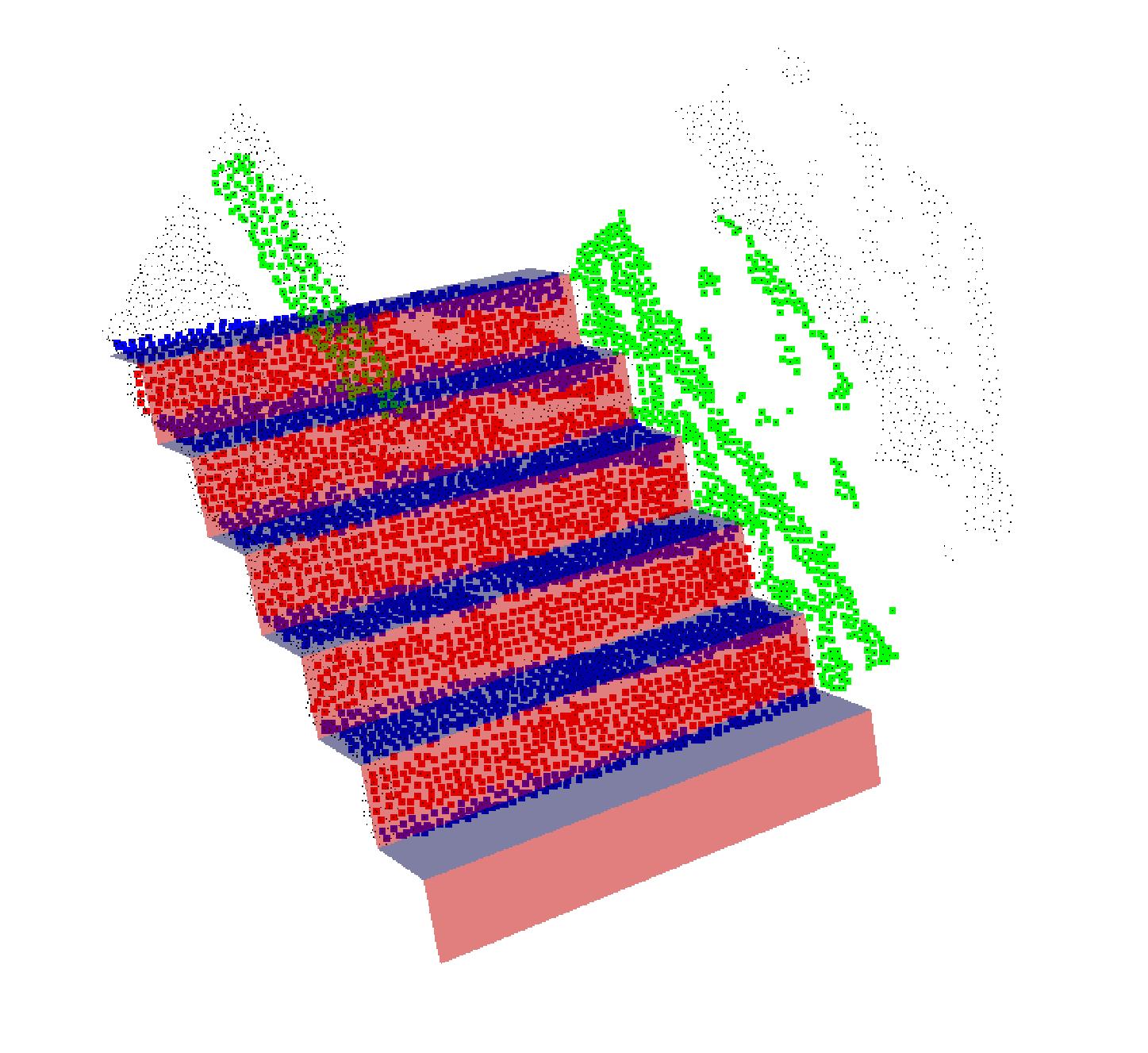}\\
                \vspace{0.35cm}
                \includegraphics[width=1.17\linewidth]{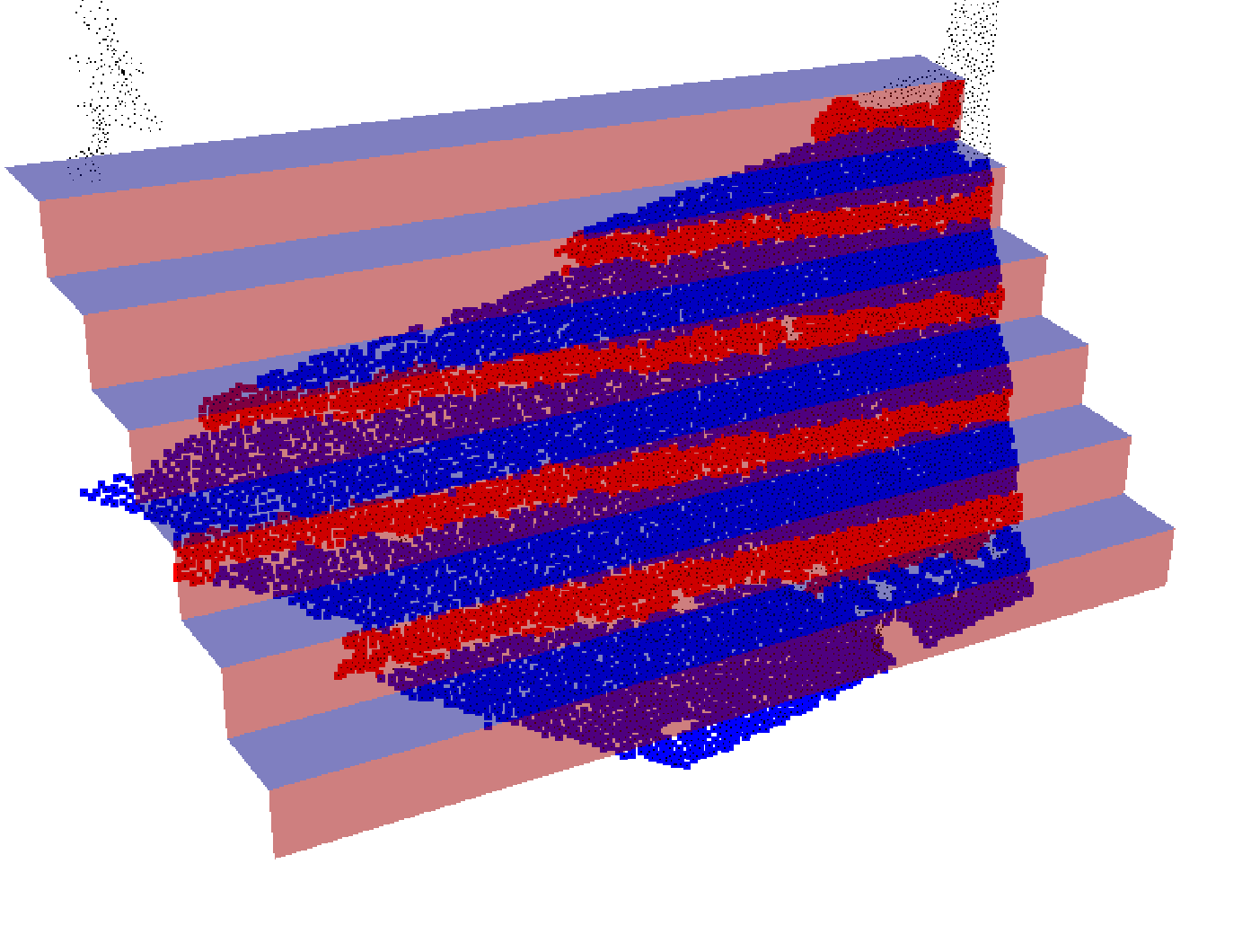}\\
			\vspace{0.07cm}
		\end{minipage}%
	}%
	\subfigure[Ours]{
		\begin{minipage}[t]{0.15\linewidth}
			\raggedright
			\includegraphics[width=1.2\linewidth]{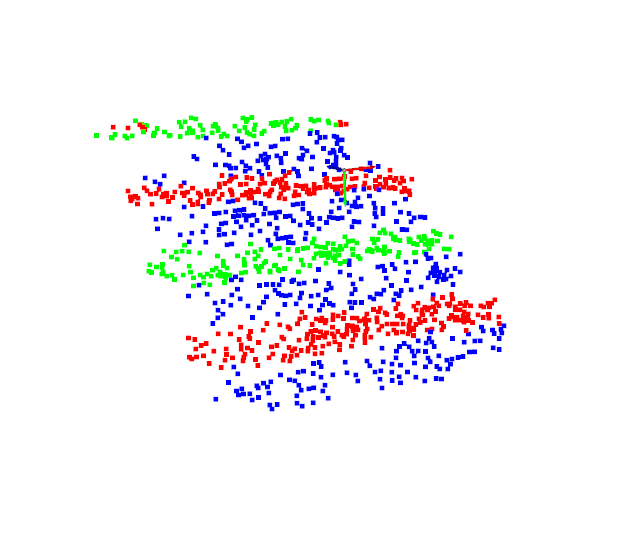}\\
                \vspace{0.4cm}
                \includegraphics[width=1.2\linewidth]{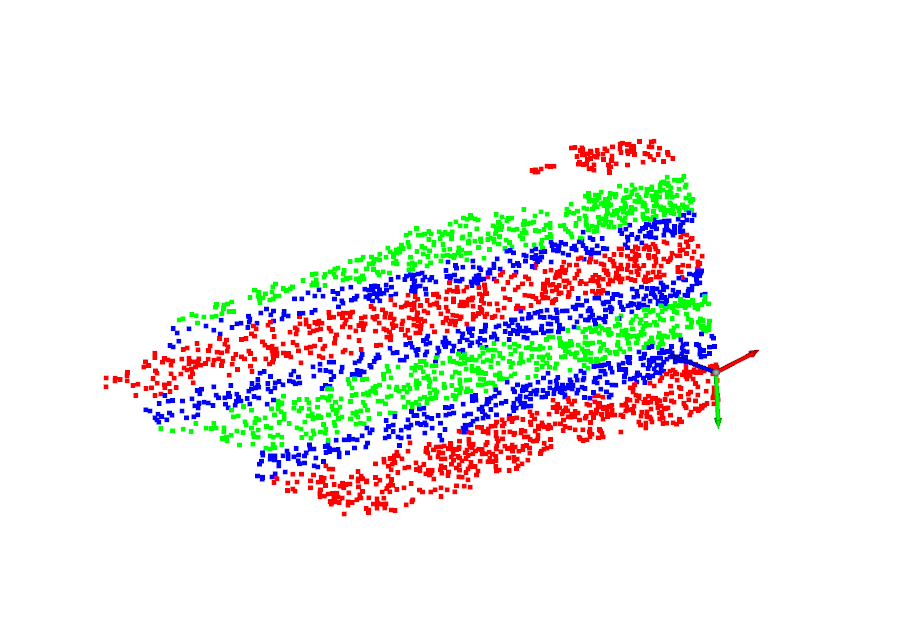}\\
			\vspace{0.13cm}
		\end{minipage}%
	}%
 	\subfigure[The raw input]{
		\begin{minipage}[t]{0.15\linewidth}
			\raggedright
			\includegraphics[width=1.2\linewidth]{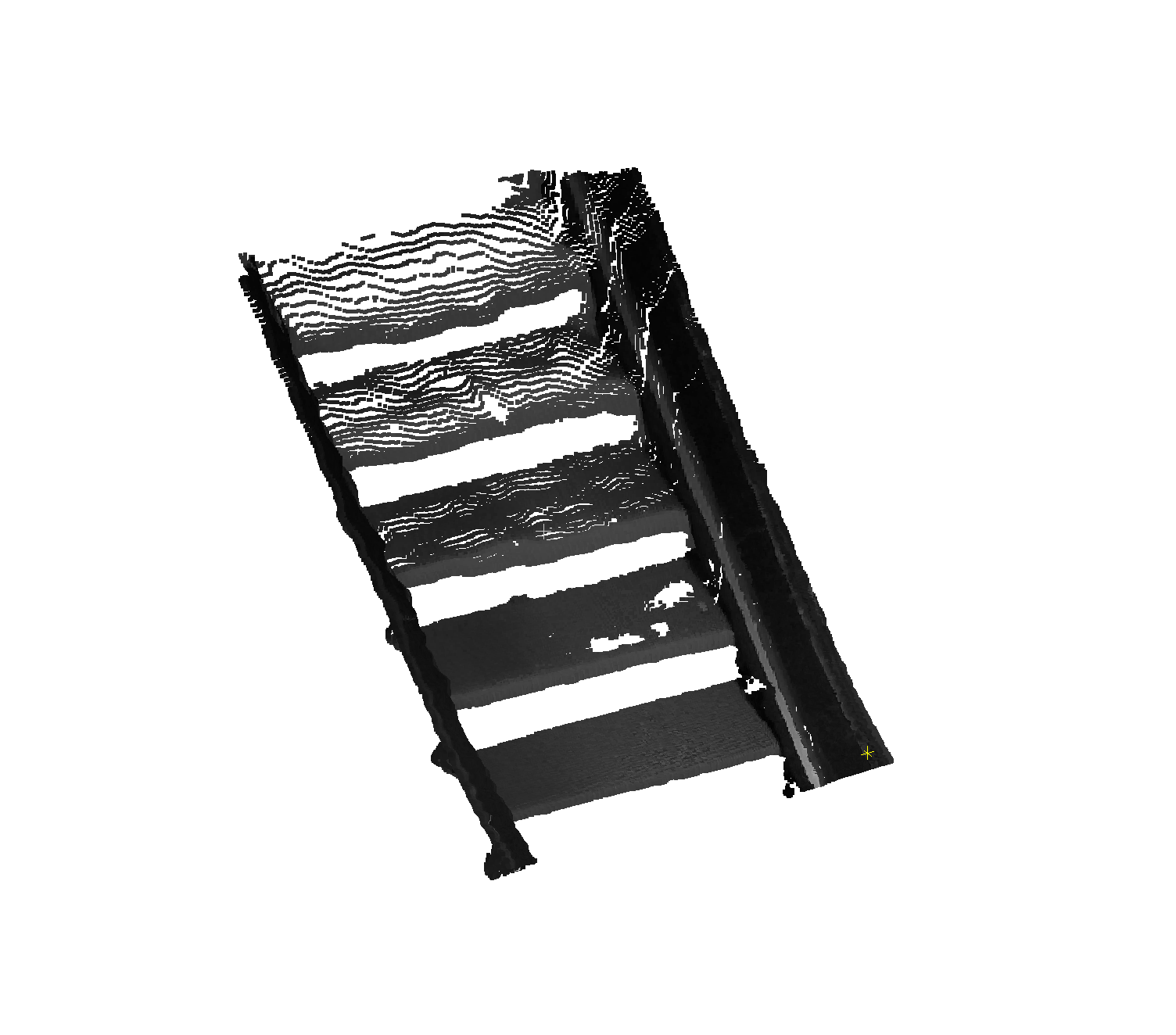}\\
                \vspace{0.02cm}
                \includegraphics[width=1.2\linewidth]{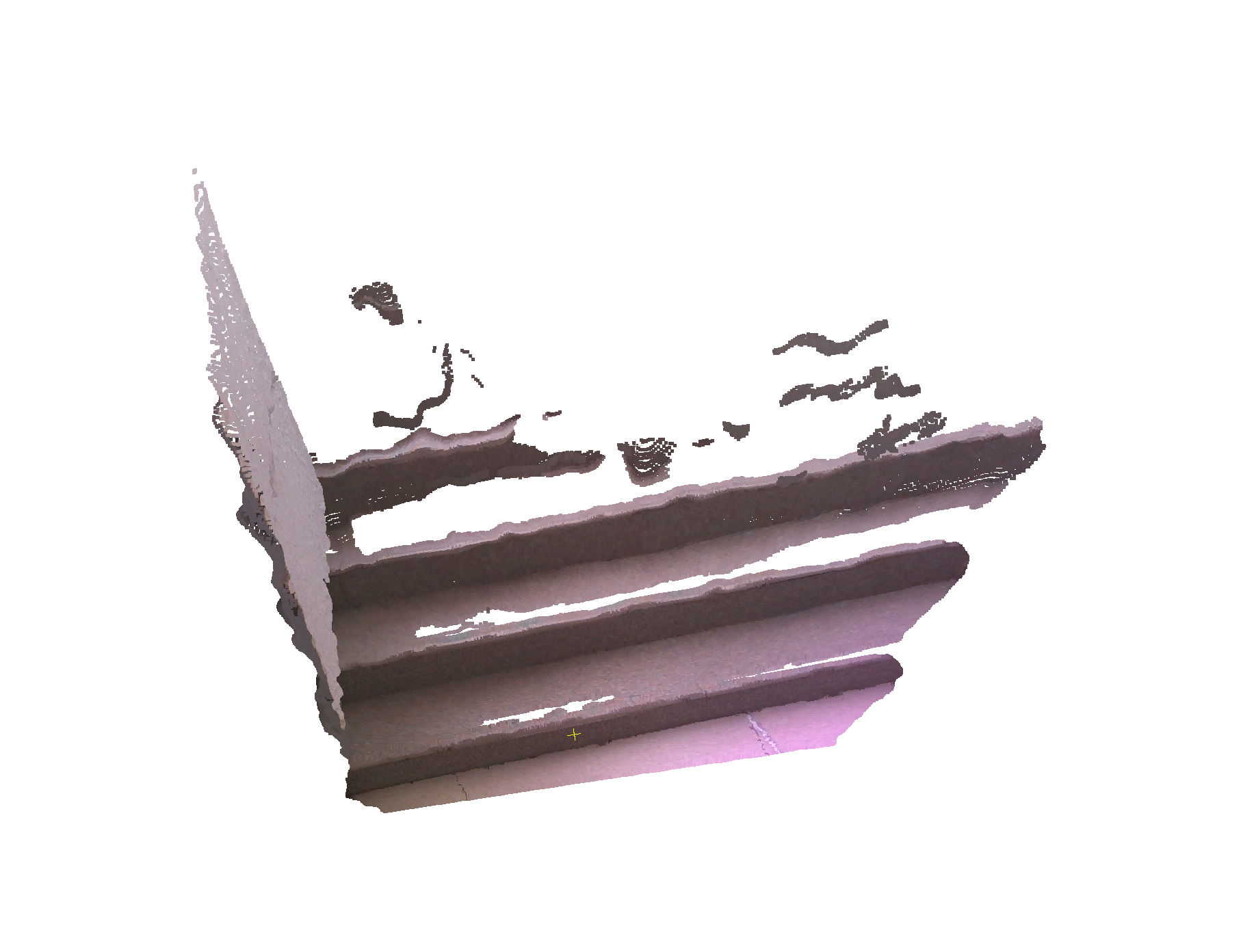}\\
			\vspace{0.36cm}
		\end{minipage}%
	}%
 	\subfigure[Thomas]{
		\begin{minipage}[t]{0.15\linewidth}
			\raggedright
			\includegraphics[width=1.2\linewidth]{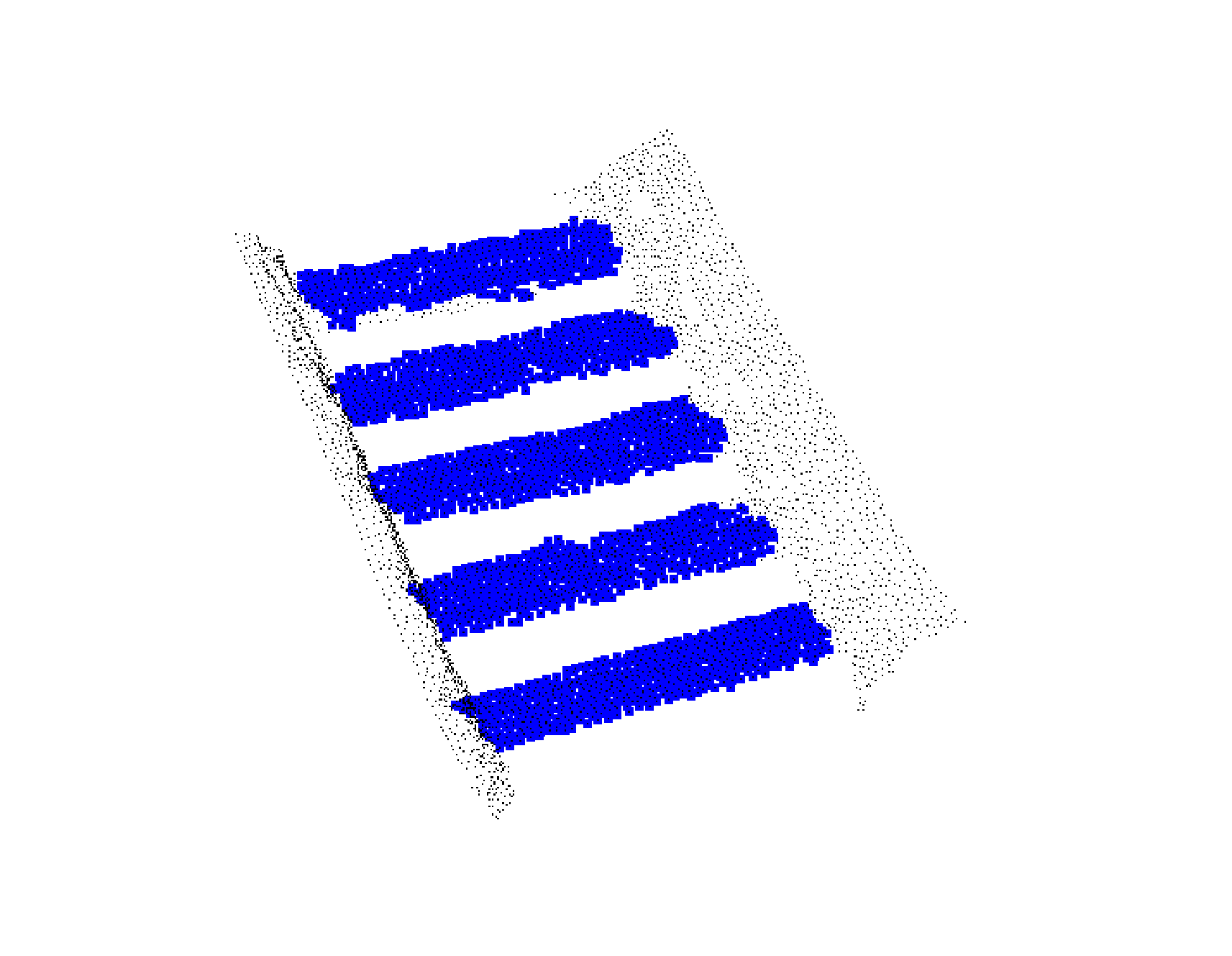}\\
                \vspace{0.08cm}
                \includegraphics[width=1.2\linewidth]{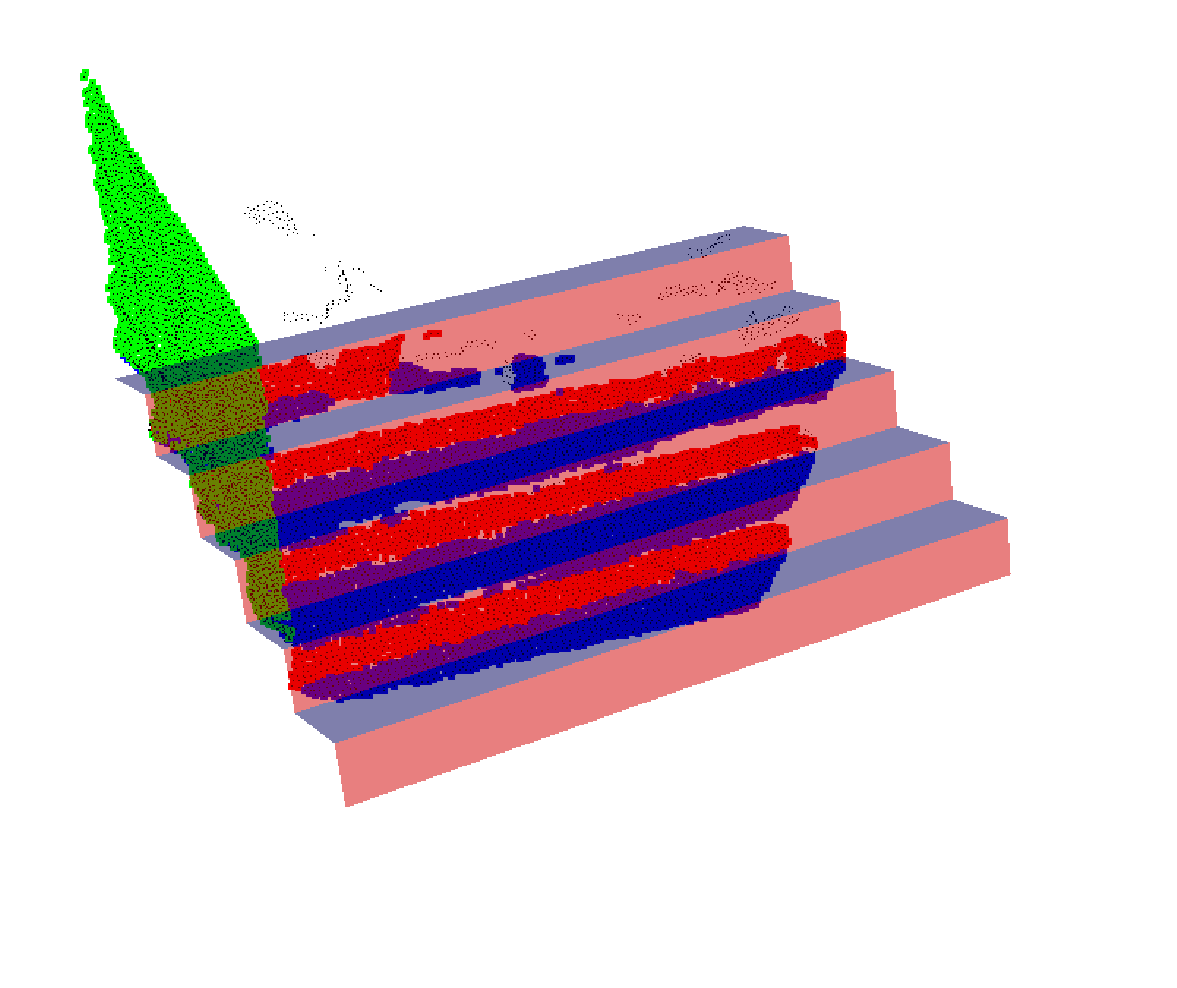}\\
			\vspace{0.07cm}
		\end{minipage}%
	}%
 	\subfigure[Ours]{
		\begin{minipage}[t]{0.15\linewidth}
			\raggedright
			\includegraphics[width=1.2\linewidth]{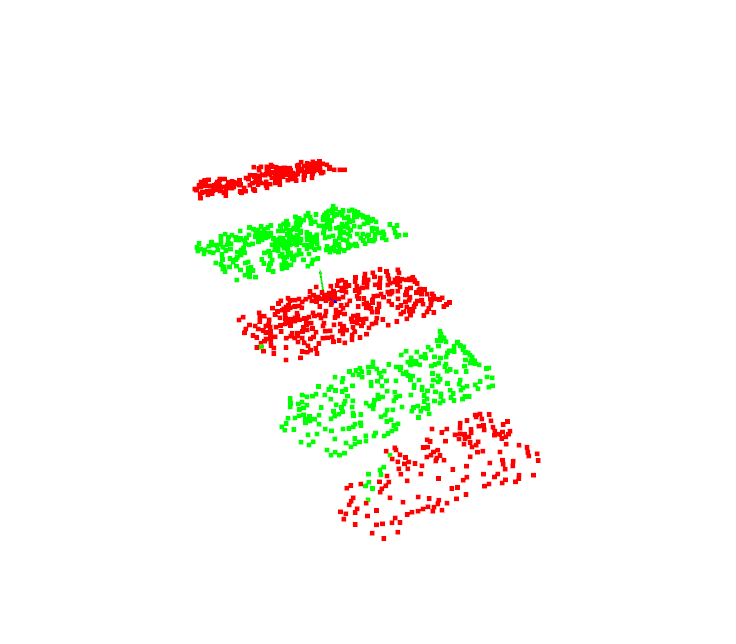}\\
                \vspace{0.02cm}
                \includegraphics[width=1.2\linewidth]{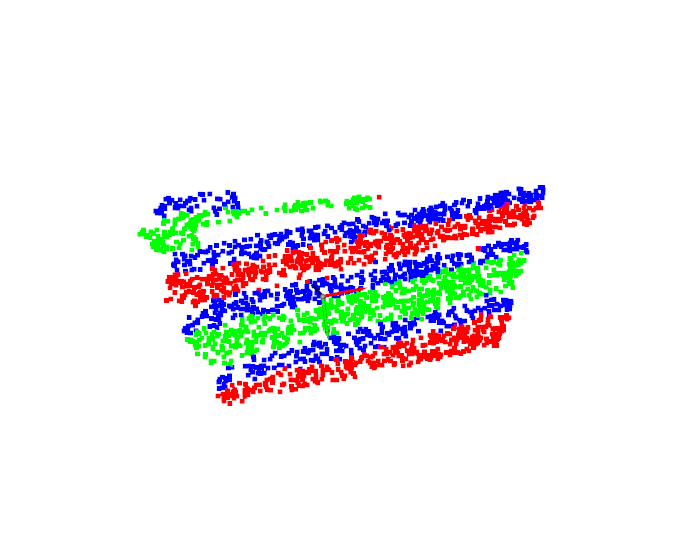}\\
			\vspace{0.27cm}
		\end{minipage}%
	}%
	\centering
	\caption{The visual results from our work and the benchmark, with the upper representing a test on the generalization set and the lower representing a test on the validation set.}
	\vspace{-0.2cm}
	\label{fig:c}
\end{figure*}

In our experiments on the generalization set, there are increasing errors in the metrics compared to those on the validation set. This can be attributed to the inherent nature of neural networks, which learn the mean and variance of features from training sets\cite{Shorten2019}. The rise in FP and FN during generalization testing is interpretable. Almost all the FP occurs in the detection of the upper step of the stair during testing. If a step is ambiguous or has a small area, we labeled it a negative sample. However, the network that is enhanced through data augmentations tends to make positive predictions for such cases, leading to an increase in FP.

In short, the data augmentations, CSCELoss, and measurement correction have demonstrated their effectiveness, resulting in acceptable performance during generalization testing.

\subsection{Comparison Experiments}

We do a comparison test on other paperwork based on the point cloud. Due to the absence of open-source codes and datasets for many experiments, we choose \cite{r10} as the benchmark and conduct comparative experiments on our dataset under the best possible adjustments. The results are presented in Table \ref{tab:CE}.

\begin{table}[!t]
\centering
\caption{Comparison Experiments}
\renewcommand\arraystretch{1.0}
\resizebox{0.5\textwidth}{!}{
\begin{tabular}{lccccc}
\toprule 
\begin{tabular}[c]{@{}c@{}}\makecell[c]{Methods}\end{tabular} &
\begin{tabular}[c]{@{}c@{}} DE(m) \end{tabular}&
\begin{tabular}[c]{@{}c@{}} HE(m) \end{tabular}&
\begin{tabular}[c]{@{}c@{}} FP(\%) \end{tabular}&
\begin{tabular}[c]{@{}c@{}} FN(\%) \end{tabular}&
\begin{tabular}[c]{@{}c@{}} TC(sec) \end{tabular} \\
\midrule 

Ours              &0.074         &\textbf{0.023}&\textbf{23.1} &\textbf{2.6}   &\textbf{$\leq$0.2}\\
Thomas\cite{r10}  &\textbf{0.056}&0.112         &48.7          &30.8           &0.706\\
Gen-Ours          &0.06          &\textbf{0.029}&\textbf{40}   &\textbf{14}    &\textbf{$\leq$0.2}\\
Gen-Thomas        &\textbf{0.035}&0.082         &44            &38             &0.726\\

\bottomrule
\vspace{-8mm}
\end{tabular}
}
\label{tab:CE} 
\end{table}

The experimental results show, that our framework demonstrates better performance in the HE, FP, FN and generalization ability at lower fixed runtime compared to the benchmark. We provide visual results to demonstrate the comparison, as shown in Fig. \ref{fig:c}. From the illustration, \cite{r10}'s method sometimes predicts a riser on the floor, and deviations with respect to the staircase pose, occasionally failing to detect. This is especially noticeable for staircases without risers, where the performance is subpar. Our approach, on the other hand, tends to make errors in predicting two treads within a single real tread, primarily occurring in elevated treads. Nevertheless, it's worth noting that both methods demonstrate effective noise suppression for outliers present on the staircase. However, our system performs relatively worse in depth error of both the validation set and the generalization set compared to the benchmark. The challenging normal estimation of risers makes evaluations worse resulting in a higher error. 

To summarize, in the test of HE, FP, FN, computing time, and generalization ability, our system performs better compared to the benchmark.


\section{Conclusion}
In this paper, we presented an end-to-end staircases instance segmentation and
modeling framework. We introduced a series of data augmentations to enhance the training of the fundamental network. We proposed a curvature suppression cross-entropy(CSCE) to reduce the ambiguity of prediction on
the boundary between traversable and non-traversable regions. We collected a dataset about staircases and introduced evaluation criteria. Compared to previous work of \cite{r10}, Our research attains a harmonious equilibrium between real-time computational efficiency and modeling precision. In future work, we would like to implement the algorithm in the legged robot to facilitate the traversal of staircases.


\bibliographystyle{IEEEtran}
\bibliography{IEEEabrv,Ref}


\end{document}